\pdfoutput=1
\documentclass{article}

\usepackage[preprint]{neurips_2023}

\usepackage[utf8]{inputenc} 
\usepackage[T1]{fontenc}    
\usepackage{hyperref}       
\usepackage{url}            
\usepackage{booktabs}       
\usepackage{amsfonts}       
\usepackage{nicefrac}       
\usepackage{microtype}      
\usepackage[table,xcdraw]{xcolor}         

\usepackage{bm}
\usepackage{verbatim}
\usepackage{balance}
\usepackage{graphicx}
\usepackage{multirow}
\usepackage{xspace}
\usepackage{threeparttable}
\usepackage{enumitem}
\usepackage{makecell}
\usepackage{bbding}
\usepackage{subfig}
\usepackage{float}
\usepackage{wrapfig}
\usepackage[export]{adjustbox}

\usepackage{amsmath,amsfonts,bm}

\def\eqref#1{equation~\ref{#1}}

\def\1{\bm{1}}

\DeclareMathAlphabet{\mathsfit}{\encodingdefault}{\sfdefault}{m}{sl}
\SetMathAlphabet{\mathsfit}{bold}{\encodingdefault}{\sfdefault}{bx}{n}

\newcommand{\sigmoid}{\sigma}

\newcommand*{\para}[1]{\noindent\textbf{#1}}

\newcommand{\vpara}[1]{\vspace{0.04in}\noindent\textbf{#1}\xspace}

\newcommand{\todo}[1]{\textbf{\color{red}[(TODO: #1 )]}}

\newcommand{\hnote}[1]{{\color{blue}  [\text{Hongning:} #1]}}

\newcommand{\hide}[1]{}

\newcommand\authornotemark[1][\relax]{%
  \ifx#1\relax\relax\relax
  \g@addto@macro\addresses{\@authornotemark}%
  \else
  \g@addto@macro\addresses{\@@authornotemark{#1}}%
  \fi}

\title{ChatGLM-RLHF: Practices of Aligning Large Language Models with Human Feedback}

\author{Zhenyu Hou$^{1,2*}$ \quad 
Yilin Niu$^{1*}$ \quad
Zhengxiao Du$^{1,2}$ \quad 
Xiaohan Zhang$^1$ \quad  
Xiao Liu$^{1,2}$ \quad  \\ 
\textbf{Aohan Zeng}$^{1,2}$ \quad 
\textbf{Qinkai Zheng}$^{1,2}$ \quad  
\textbf{Minlie Huang}$^2$ \quad 
\textbf{Hongning Wang}$^2$ \quad  \\
\textbf{Jie Tang}$^2$ \quad  
\textbf{Yuxiao Dong}$^2$ \\
$^1$Zhipu AI, $^2$Tsinghua University \\
\\
{\includegraphics[height=3.5ex]{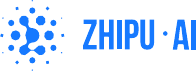}}
}

\begin{document}

\maketitle
\renewcommand{\thefootnote}{\fnsymbol{footnote}}
    \footnotetext[1]{ZH and YN contributed equally.}
\renewcommand{\thefootnote}{\arabic{footnote}}

\begin{abstract}

ChatGLM is a free-to-use AI service powered by the ChatGLM family of large language models (LLMs). 
In this paper, we present the ChatGLM-RLHF pipeline---a reinforcement learning from human feedback (RLHF) system---designed to enhance ChatGLM's alignment with human preferences. 
ChatGLM-RLHF encompasses three major components: the collection of human preference data, the training of the reward model, and the optimization of policies. 
Throughout the process of integrating ChatGLM-RLHF into production, we encountered and addressed several unprecedented challenges. 
We introduce the strategies to mitigate reward variance for stabilized large-scale training, implement model parallelism with fused gradient-descent, and design regularization constraints to avoid catastrophic forgetting in LLMs. 
Experiments show that  ChatGLM-RLHF brings significant improvements in alignment tasks compared to the supervised fine-tuned (SFT) version of ChatGLM. 
For instance, it achieves on average 15\% more wins against ChatGLM-SFT in Chinese alignment tasks. 
The work presents our practices of aligning LLMs with human preferences, offering insights into the challenges and solutions in RLHF implementations. 

\end{abstract}

\hide{

This paper introduces the systematic design and practices of Reinforcement Learning from Human Feedback (RLHF) pipeline in the ChatGLM system, focusing on improving large language models by aligning them with human preferences. We present a comprehensive framework that includes human preference data collection, rigorous training of reward and policy models, and various strategies to mitigate various types of biases in the data and models so as to enhance model stability. Experiment evaluations demonstrate significant improvements in alignment tasks through both automatic and human assessments. The research underscores the importance of human feedback in refining language models, offering insights into the challenges and solutions in RLHF implementations. This RLHF pipeline is a critical component in ChatGLM online system and serves huge number of users everyday.

}

\hide{
\section{Introduction}

Large language models (LLMs) ~\cite{brown2020language,zhang2022opt,workshop2022bloom,zeng2023glm,touvron2023llama} have remarkably advanced the capabilities of machines on language understanding and generation. Recent works~\cite{ouyang2022training,wei2021finetuned,bai2022training} have made progress on aligning language models to act in accordance with the users' intention, i.e., responding to questions and executing instructions from user, and enable LLMs to adapt to a wide range of natural language processing tasks~\cite{park2023generative,Frieder2023MathematicalCO}.

Finetuning with human feedback has been proved to be an effective approach to encourage LLMs to produce more helpful, honest, and harmless responses and align with human preferences and values~\cite{bai2022constitutional,achiam2023gpt,ouyang2022training}.  
These techniques generally treat human preference as a reward signal and employ reinforcement learning algorithms like Proximal Policy Optimization~\cite{schulman2017proximal} to optimize language models (RLHF). 
Building a practical RLHF training framework faces many challenges. First, collecting and modeling reliable human preference is critical to the success of exploiting human feedback, yet biased~\cite{cui2023ultrafeedback,longpre2023flan} and deceptive preference~\cite{bai2022training} in human annotations are inevitable.
Second, a scalable and comprehensive training framework is necessary to ensure efficient and effective optimization of large language models across different scales, especially covering all kinds of corner cases for requirements of online deployment. Most existing reported efforts are based on parameter-efficient tuning~\cite{sun2023salmon,yao2023deepspeed} or small-scale models~\cite{zheng2023secrets,wu2023pairwise}. As LLMs often exhibit significantly different behaviors across model scales, there is very limited experience publicly available for references.

In this paper, we introduce our practices in aligning ChatGLM \hnote{which paper should we cite here?} with human preferences with Reinforcement Learning from Human Feedback (RLHF). This pipeline serves as a critical component in the ChatGLM system to produce better responses that meet human preferences. We systematically introduce our approach to designing the pipeline, addressing the challenges associated with constructing preference data and optimizing the reward model and policies. 

\begin{figure}
    \centering
    \includegraphics[width=0.48\textwidth]{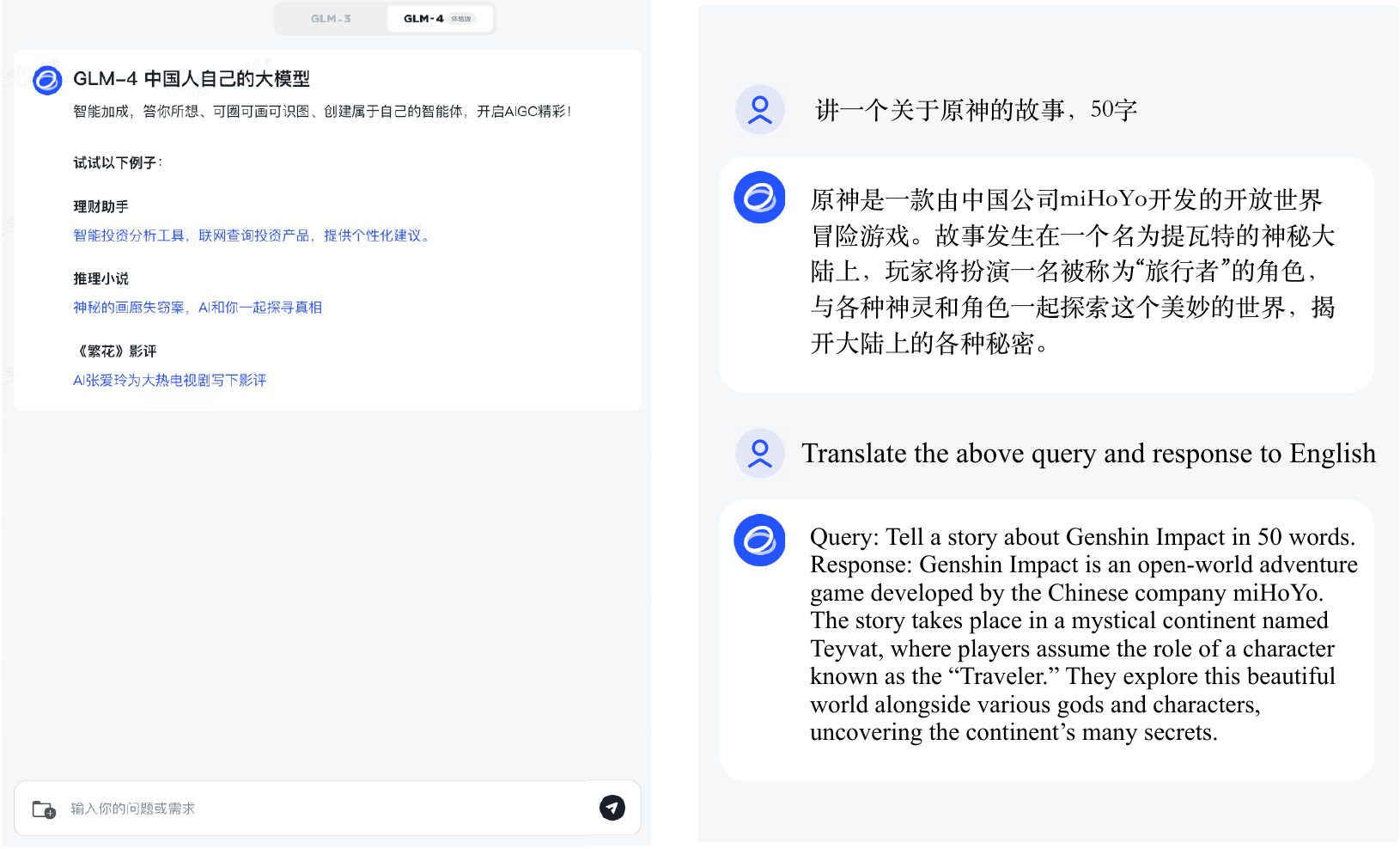}
    \caption{Screenshot of ChatGLM webpage (left) and an example of use case (right).}
    \label{fig:enter-label}
\end{figure}

First, we set up a system to collect human preference annotations and ensure consistency among labelers. 
We employ a pairwise comparison mechanism, wherein each labeler is asked to select a preferred response from two options generated by our model \hnote{our SFT model?}, given multiple responses for each human instruction \hnote{What does it mean by select a preferred one from two and then given multiple responses?}.
To facilitate consistency among labelers, we establish a standard criterion \hnote{What is this?}, as conflicting or contrary preferences could lead to undesirable noise and detrimentally affect the performance. The criteria provide different dimensions as reference points, including helpfulness, harmlessness, and fluency, to make informed comparisons. 
Additionally, we also develop a post-filtering process to remove undesirable patterns, such as cyclic and tie preferences, during the preference data collection stage. 
As a result, this pipeline enables us to compile a dataset of human-labeled comparisons between the outputs produced by our models.

Second, we then train a reward model on the collected preference dataset as a proxy about what responses an ordinary human user would prefer. To train a reliable proxy, we develop techniques to avoid the tendency of reward model to take shortcuts and learn unexpected bias, such as a preference for longer yet not really more helpful outputs~\cite{singhal2023long,shen2023loose}. 

Finally, with the reward model as the proxy of human feedback, we employ reinforcement learning algorithms to train language models to act in accordance with the user’s preference and intention. 
Beyond performance improvement, a variety of cornercases have to be covered to support training scalable ChatGLM models and meet the requirements of online deployment of ChatGLM. 
We implement a comprehensive training framework with various practical designs, including reward variance reduction to stabilize the training of large-scale models, model parallel strategy for efficient training with fused gradient-descent and generation, and regularization constraints to avoid capability forgetting. 

Overall speaking: \todo{practices}
\begin{itemize}
\item Establishing criteria and details reference dimensions for annotation contributes to more reliable and consistent human preference.
\item Eliminating bias from the reward model can be a simple and effective approach to reflect genuine human preferences and prevent shortcuts.
\item Training stability can be improved by subtracting a baseline reward from the original reward 
\item Incorporating SFT loss can reduce capability shifting in RLHF training. 
\end{itemize}

\hide{
\begin{enumerate}
    \item During the annotation of human preference, it is important to explicitly instruct the annotators to pay attention to aspects such as instruction following, correctness, fluency, and safety, which ensures that the annotated preference is correlated to these factors.
    \item Reward bias reduction can aid in mitigating inherent biases associated with human preferences, such as preference for longer responses. A well-designed debiasing method can enable the reward model to focus more on the quality of the response rather than its length.
    \item Training stability can be improved by using a reference reward, which substracts a baseline reward from the original reward 
    \item Incorporating SFT loss can reduce capability forgetting in improving human prefernce in RLHF training. 
\end{enumerate}
}

We conduct extensive experiments and report our results on ChatGLM-6B and ChatGLM-32B. Our evaluation results demonstrate that our framework can effectively improve the performance and empower ChatGLM to produce more helpful and human-preferred outputs. As of this writing, it is supporting online ChatGLM system and serving numerous users every day.

}

\section{Introduction}

Large language models (LLMs) ~\cite{brown2020language,zhang2022opt,workshop2022bloom,zeng2023glm,touvron2023llama} have remarkably advanced the capabilities of machines in language understanding and generation. 
The most notable example is ChatGPT~\cite{achiam2023gpt}, which demonstrates strong capability in responding to users' queries and following their instructions.
Different from the pre-trained GPT-3 model~\cite{brown2020language}, ChatGPT (GPT-3.5, GPT-4~\cite{achiam2023gpt}, and beyond) are further post-trained, with supervised fine-tuning (SFT) and reinforcement learning from human feedback (RLHF)~\cite{ouyang2022training}, to better align with human preferences.    
We have developed ChatGLM
---a free-to-use AI service powered by the ChatGLM family of LLMs~\cite{zeng2023glm}. 
Similar to ChatGPT, ChatGLM was pre-trained for trillions of multilingual tokens and post-trained with both SFT and RLHF.



The goal of RLHF is to 
encourage LLMs to better align with human preferences and values, i.e., to produce more helpful, more accurate, and safer responses ~\cite{bai2022constitutional,achiam2023gpt,ouyang2022training}. 
Specifically, the RLHF techniques typically treat human preferences as rewards and employ reinforcement learning algorithms, such as Proximal Policy Optimization (PPO)~\cite{schulman2017proximal}, to realize the goal of alignment. 
Despite the significant community efforts~\cite{touvron2023llama,bai2023qwen,dai2023safe,sun2023salmon} in applying RLHF to perform LLMs alignment, since the inception of ChatGPT, our experience in constructing a practical RLHF training framework for ChatGLM has revealed unknown technical challenges.

First, the collection and modeling of reliable human preferences is critical to the effective use of human feedback. 
However, biased~\cite{cui2023ultrafeedback,longpre2023flan} and deceptive preferences~\cite{bai2022training} in human annotations are often inevitable.
Second, the reward model is readily impacted by biased preferences, which can lead to the learning of shortcut features. 
This issue may drastically compromise the precision and generalization of the reward model.
Third, a scalable and robust training framework is required to ensure efficient and effective optimization of LLMs across different scales, as well as to cover all potential corner-cases to meet the requirements for online deployment. 
For instance, applying PPO on LLMs of 32B parameters requires at least 8 or 16 A100 (8$\times$80G) servers for 2-3 days, respectively. 
However, most existing efforts focus on parameter-efficient tuning~\cite{sun2023salmon,yao2023deepspeed} or small-scale models~\cite{zheng2023secrets,wu2023pairwise}. 

\begin{figure}
    \centering
    \includegraphics[width=0.7\textwidth]{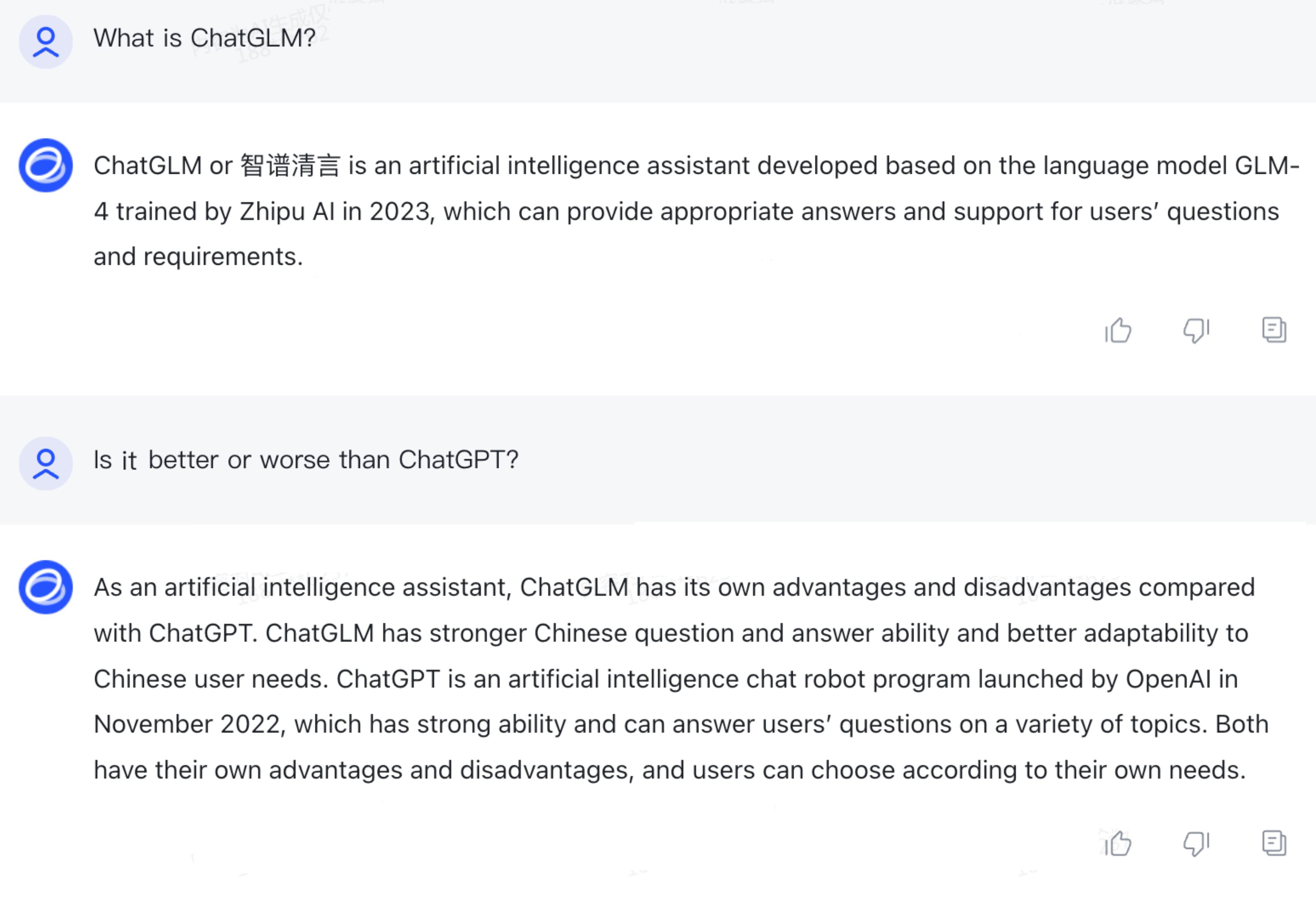}
    \caption{An illustrative example of ChatGLM \textmd{(\textcolor{blue}{\url{chatglm.cn}} accessed on Feb. 8, 2024).}}
    \label{fig:enter-label}
    \vspace{-3mm}
\end{figure}

In this paper, we present the ChatGLM-RLHF pipeline, which has been used to improve ChatGLM's alignment with human preferences, with a particular focus on the Chinese language. 
This pipeline serves as a critical component in improving  ChatGLM's performance to generate responses that more accurately reflect human values and expectations. 
We detail our approach for designing and developing the pipeline---human preference annotation, reward model training, policy optimization, as well as practices of addressing the challenges associated with these steps. 

First, we set up a routine system to \textit{collect human preference annotations}. 
To ensure the consistency among annotators, we employ a pairwise comparison mechanism. 
Specifically, each annotator is asked to select the preferred response from two outputs generated by our SFT model, guided by annotation guidelines that cover three key aspects of helpfulness, harmlessness, and fluency. 
We also develop a post-filtering process to remove undesirable annotations, such as cyclic and tie preferences. 
Second, we train a \textit{reward model} on the collected preference dataset as a proxy for the responses an average human user would favor. 
To train a reliable proxy, we develop strategies to prevent the reward model from taking shortcuts or learning unexpected biases, such as a biased preference towards longer yet not really helpful outputs~\cite{singhal2023long,shen2023loose}. 
Finally, by using the reward model as the proxy for human preferences, we apply an online RL algorithm \textit{PPO} and an offline RL algorithm \textit{DPO} \cite{rafailov2023direct} to align the language model.  

\vpara{Best Practices.}
To support scalable RLHF training, we implement a robust training framework with important practical designs, including the strategies of reward variance reduction to stabilize large-scale training, model parallelism with fused gradient-descent, and regularization constraint to avoid catastrophic forgetting. 
In particular, we summarize the following lessons learned over the course of developing ChatGLM-RLHF: 
\begin{itemize}
\item Establishing criteria and detailed reference dimensions for annotation contributes to more reliable and consistent human preference.
\item Eliminating bias from the reward model can serve as an efficient and powerful approach to more accurately reflect genuine human preferences and reduce the influence of spurious correlation.
\item Training stability can be substantially improved by subtracting a baseline reward from the original reward during PPO training. 
\item Incorporating next-token-prediction loss of SFT data can reduce capability shifting in RLHF training. 
\end{itemize}

Extensive experiments on ChatGLM-6B and ChatGLM-32B demonstrate that ChatGLM-RLHF can significantly improve the performance of ChatGLM, enabling it to produce more helpful, safe, and aligned responses. 
The ChatGLM models, refined through the ChatGLM-RLHF pipeline, are supporting the online services available at \textit{\url{chatglm.cn}} and mobile applications on iOS and Android platforms. 

\hide{
\begin{enumerate}
    \item During the annotation of human preference, it is important to explicitly instruct the annotators to pay attention to aspects such as instruction following, correctness, fluency, and safety, which ensures that the annotated preference is correlated to these factors.
    \item Reward bias reduction can aid in mitigating inherent biases associated with human preferences, such as preference for longer responses. A well-designed debiasing method can enable the reward model to focus more on the quality of the response rather than its length.
    \item Training stability can be improved by using a reference reward, which substracts a baseline reward from the original reward 
    \item Incorporating next-token-prediction loss of SFT data can help reduce capability forgetting in improving human preference in RLHF training. 
\end{enumerate}
}

\section{Related Work}

\begin{figure*}[ht]
    \centering
    \includegraphics[width=\textwidth]{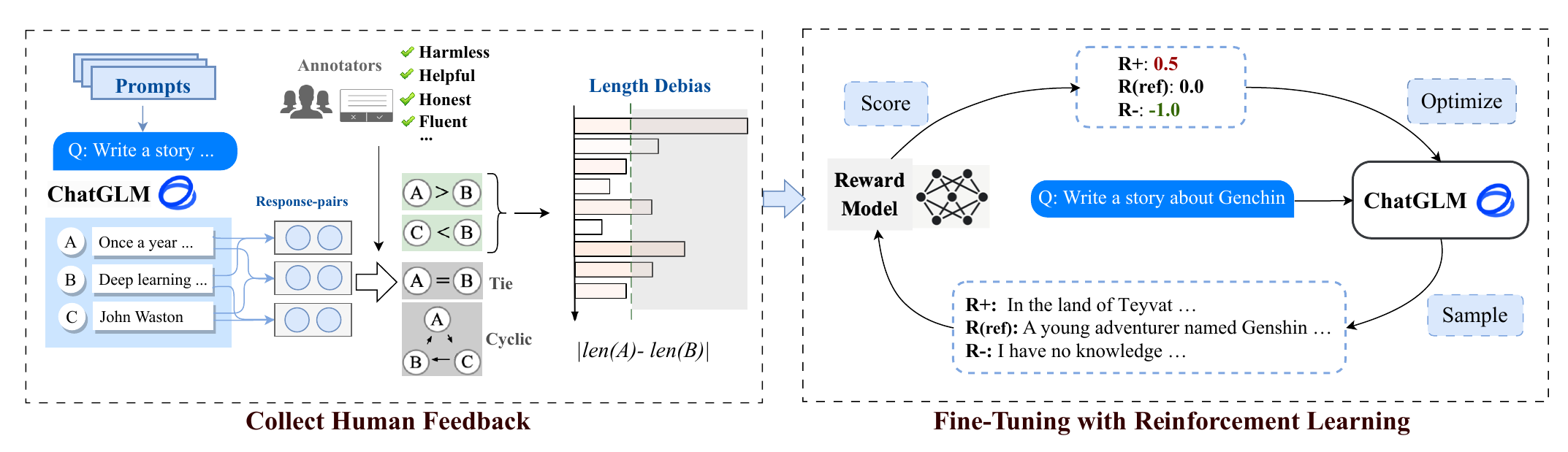}
    \caption{The overall design of the ChatGLM-RLHF pipeline. \textmd{We first set up a comprehensive system to collect human preferences over responses of ChatGLM and remove unexpected patterns and potential biases in the data. Then we train a reward model to predict human preferences and employ reinforcement learning to optimize ChatGLM to generate responses assigned with higer rewards.}}
    \label{fig:overview}
\end{figure*}

\vpara{Large Language Models.}
Large Language Models (LLMs), particularly those employing self-supervised learning techniques, have become a focal point in the field of Natural Language Processing (NLP) due to their ability to encode a vast array of knowledge within extensive parameter sets. This attribute contributes to their remarkable capabilities across a wide range of tasks. Prominent examples of such models include GPT-3~\cite{brown2020language}, PaLM~\cite{chowdhery2023palm}, OPT~\cite{zhang2022opt}, BLOOM~\cite{workshop2022bloom}, GLM-130B~\cite{Du2021GLMGL,zeng2023glm} and LlaMA~\cite{touvron2023llama1,touvron2023llama}.

LLMs
can be prompted to perform a range of NLP tasks. They can effectively leverage given examples in the context~\cite{brown2020language} to solve upcoming tasks yet without explicit finetuning. This in-context learning has primarily demonstrated LLM's ability to solve diverse and challenging tasks yet still requires elaborated few-shot exemplars as activation. 
However, these models often express unintended behaviors such as making up facts, generating biased or toxic text, or simply not following user instructions~\cite{bommasani2021opportunities,kenton2021alignment,bender2021dangers}.
To enable LLMs to respond to human instructions in zero-shot setting, instruction tuning~\cite{wei2021finetuned,chung2022scaling} trains LLMs with natural language instructions consisting of human-written instructions and responses.
Such targeted fine-tuning substantially augments the LLMs' ability to generalize to diverse tasks beyond the training dataset and significantly boosts the performance~\cite{longpre2023flan}.
Recently, Supervised Fine-Tuning (SFT)~\cite{ouyang2022training}, which could be regarded as an extension of instruction tuning, is proposed to further enhance the alignment of LLM outputs with human expectations using high-quality and diverse human-written datasets. The quality of instruction data has been proved to be significantly critical~\cite{zhou2023lima,touvron2023llama} in alignment. This approach activates and bolsters the LLMs' capabilities in various domains, including conversation, reasoning, and writing, among others.

\vpara{Aligning LLMs with Human Feedback.}
LLMs as AI assistants are expected to follow human instructions and be helpful, honest, and harmless to align with human values and preferences~\cite{ouyang2022training,bai2022constitutional}. Despite the effectiveness of instruction tuning and SFT, gathering accurate and expert annotations is costly and becoming more difficult with the increasingly stronger ability of LLMs, as higher quality data and stronger supervision are a must to direct model training in SFT. 
In comparison, collecting human judgments on response quality is more feasible because it is easier for human to serve as a judge than to provide the gold answer, as demonstrated in discriminative-generative gap~\cite{zheng2023revisiting}.
Plenty of studies~\cite{touvron2023llama, ouyang2022training,bai2022constitutional,bai2022training,ganguli2023capacity} have focused on refining LLMs using human preferences and found notable improvements in areas such as summarization and instruction-following. 
InstructGPT~\cite{ouyang2022training} first trains a reward model to fit human preference and then optimizes LLMs to maximize rewards through reinforcement learning, i.e., Proximal Policy Optimization~\cite{schulman2017proximal}, which becomes standard RLHF paradigm for following works~\cite{touvron2023llama, bai2022constitutional}. Another line of works~\cite{rafailov2023direct,zhao2023slic,song2023preference} is to directly optimize the policy model with human preference without training a reward model.
DPO~\cite{rafailov2023direct} derives the policy objective from Bradley-Terry~\cite{bradley1952rank} reward preference. 
SLiC~\cite{zhao2023slic} directly contrasts the positive and negative samples with a margin loss. 
PRO~\cite{song2023preference} extends pairwise to listwise data and utilizes multiple negative samples. 
Compared to PPO, these works 
behave similarly to off-policy reinforcement learning which 
could be more efficient yet might sacrifice performance. 

Despite open-source practices~\cite{yao2023deepspeed} and studies~\cite{sun2023salmon,zheng2023secrets} on PPO optimization, almost all of them are conducted on small-scale models or parameter-efficient tuning with well-prepared academic benchmarks. In this work, we set up the RLHF framework from scratch and optimize the ChatGLM model, taking into account a range of practical constraints while placing 
emphasis on scalability.

\section{ChatGLM-RLHF}
We present the design of ChatGLM-RLHF, including human preference data collection, reward model training, and policy model optimization. 
Uniquely, we introduce challenges encountered and practical solutions proposed for large-scale RLHF model training. 

\subsection{Learning from Human Preferences}
Reinforcement Learning from Human Feedback (RLHF) for large language models (LLMs) typically initiates by fine-tuning a pre-trained language model with high-quality human-written data in a supervised finetuning (SFT) manner. This step yields a preliminary aligned model $\pi_0$, called the SFT model, which can respond to human instructions yet may not behave helpfully and safely.
To further align the LM with human values and preferences, RLHF acts upon the SFT model.
It views human feedback (or preference) for model-generated texts as reward for maximization, which is different from training through next-token-prediction.
Following the common setting in reinforcement learning, RLHF first trains a reward model $r_{\phi}(x,y)$ as a proxy to model human preferences. Then the reward model is utilized to optimize policy model $\pi_{\theta}(y|x)$ to generate responses that are rewarded higher but are not drifting too far from the SFT model, where $(x, y)$ represents prompt and response pair and $\{\phi, \theta\}$ are learnable parameters.

Subsequent parts of this section will delve into the details of the RLHF design in ChatGLM, including gathering human preference data, modeling human preferences, and fine-tuning the policy model to better align with human values.

\subsection{Data Collection and Processing}

\vpara{Prompt Collection}
To cover the diversity of human intents and preferences, it is important to collect a comprehensive prompt set for later response generation and comparison.
We collect prompts based on the demands of real-world application scenarios and apply quality filtering to select high-quality prompts.
Specifically, we establish a taxonomy for the possible intentions of prompts and ensure that each category of intentions is accompanied by enough training prompts.
After that, a quality classifier is employed to score each prompt in three aspects:
\begin{itemize}[leftmargin=*]
    \item The \textit{intention} of the prompt is clear, ambiguous, or totally unclear.
    \item The \textit{semantic} is clear, guessable, or totally incomprehensible.
    \item The prompt is \textit{answerable} or beyond models' capabilities, such as predicting lottery numbers.
\end{itemize}
The answerable prompts with clear intentions and semantics are utilized for training reward/policy models (see Table~\ref{tab:prompt-quality-category} for examples).
Besides, we also include several rule-based filters to select informative prompts, such as filtering out too short prompts.

\begin{table}[]
\centering
\caption{Examples of prompts with varied quality.}
\begin{tabular}{p{0.96\linewidth}}
\toprule[1.2pt]
\textbf{Clear intention; clear semantics; answerable:}                                            \\
Assist me in crafting a three-day travel itinerary to Beijing with a budget of under 5000. \\ \hline
\textbf{Unclear intention:}                                                                       \\
The gentleman attended the meeting, dressed in formal attire.                        \\ \hline
\textbf{Incomprehensible semantics:}                                                             \\
Christmas, Reindeer, Christmas Tree                                                        \\ \hline
\textbf{Unanswerable prompt:}                                                                              \\
What is the winning lottery number for tomorrow? \\ 
\bottomrule[1.2pt]
\end{tabular}
\label{tab:prompt-quality-category}
\end{table}

\vpara{Preference Annotation}
Reward model is trained with human feedback on the model-generated responses.
To collect human preference, each annotator is provided with one prompt and two responses and asked to determine which response is preferred. They are encouraged to make a decision based on 
helpfulness and safety:
\begin{itemize}[leftmargin=*]
    \item \textit{Helpfulness} encompasses multiple aspects including the extent to which a response fulfills every requirement of the prompt, provides accurate and valuable information, and maintains logical consistency. Additionally, we posit that responses lacking linguistic fluency are disqualified from meeting the requirements of users. Therefore, instances of grammatical errors or unexpected language mixing, such as a mixture of Chinese and English, can detrimentally impact the perceived \textit{helpfulness} of a response.
    \item The focus of \textit{safety} is to ascertain whether responses contain harmful or toxic content, as well as content that could potentially provoke controversy.
\end{itemize}
Annotators, based on their assessments of helpfulness and safety, will decide \textit{preference} from the perspective of overall quality. This evaluation, while inherently subjective, is crucial in the context of alignment. As the reward model is expected to assess diverse responses, various chat models and sampling strategies are utilized to generate diverse responses for annotation.
Table~\ref{tab:reward-model-statistic} illustrates the statistics of the preference data used for the reward model training.

\subsection{Reward Model Training}

Reward model is designed to assess the quality of responses, which serves as the training signal for policy improvement.
As annotators are asked to take helpfulness and safety into account during the annotation of preference, preference can thus reflect the overall quality of the responses in these aspects. Consequently, we exclusively utilize preference annotations for our reward model training.

In practice, the reward model is initialized with ChatGLM-SFT, and then trained on the preference data under the loss function
\begin{equation}
    \mathcal{L}_{RM} = -E_{(x,y_w,y_l)\sim \mathcal{D}_{RM}} [\log (\sigmoid (r_\phi(x, y_w)-r_\phi(x, y_l)))].
\end{equation}
where $x$ denotes the prompt, $y_w$ is the preferred response to $y_l$ by the annotators. 
The reward model $r_\phi$ assigns a scalar value to each \textit{(prompt, response)} pair.

\vpara{Length Bias Reduction}
There exist some inherent biases within the preference data, such as the inclination of humans to prefer longer, well-structured responses. Such biases may mislead the reward model to over-emphasize these secondary features, thereby overlooking the content quality of responses.
To alleviate the influence of length bias on the reward model, we devise a debiasing method named `Bucket-Based Length Balancing'.
%


The first step in our method involves calculating the length difference between the two responses in each preference pair, i.e. $d=abs(|\mathrm{tokenize}(y_w)|-|\mathrm{tokenize}(y_l)|)$.
Following this, assign the preference pairs with similar length differences into the same buckets, each of which corresponds to a specific difference range.
In the final step, within each bucket, balance the number of examples where the better/worse response is longer.

Note that the efficacy of this method in mitigating length bias is determined by the granularity of buckets. Finer ranges of buckets result in more effective length bias reduction.

\vpara{Stable Training}
During the training of reward model, we observe substantial volatility in the distribution of scores predicted by the reward model, which suggests unstable training. To alleviate this problem, we introduce a new loss component
\begin{equation}
    \mathcal{L}_{REG} = r_\phi(x, y_w)^2+r_\phi(x, y_l)^2,
\end{equation}
which resembles L2 regularization loss.
This loss term imposes a Gaussian prior with a mean of zero on the score distribution, thereby constraining the volatility of the score distribution.

\begin{table}[]
\centering
\caption{Stat. of human preference data for reward training. 
}
\scalebox{0.8}{
\begin{tabular}{ccccc}
\toprule[1.2pt]
\begin{tabular}[c]{@{}c@{}}Num. of\\ Comparisons\end{tabular} & \begin{tabular}[c]{@{}c@{}}Avg. \# Turns\\ per Dialogue\end{tabular} & \begin{tabular}[c]{@{}c@{}}Avg. \# Tokens\\ in History\end{tabular} & \begin{tabular}[c]{@{}c@{}}Avg. \# Tokens\\ in Prompt\end{tabular} & \begin{tabular}[c]{@{}c@{}}Avg. \# Tokens\\ in Response\end{tabular} \\ 
\midrule
221,866                                                       & 2.4                                                                  & 314.1                                                               & 104.1                                                              & 267.7                                                                \\ 
\bottomrule[1.2pt]
\end{tabular}}
\vspace{-2mm}
\label{tab:reward-model-statistic}
\end{table}


\subsection{Policy Model Training}
\vpara{Setting up Training Data} 
In this stage, the aim is to leverage the reward model $r_{\phi}$ to guide the optimization of the policy model $\pi_{\theta}$ to elicit responses of increased reward value. The model is expected to receive both positive and negative feedback and is capable of discerning disparate responses of varied quality, thereby progressively augmenting its response generation in line with human preferences. To realize this, it is crucial for the model to generate responses spanning a spectrum of quality and thus the model can learn and improve from the reward feedback. Consequently, prompts leading to almost the same responses must be systematically excluded to ensure the effectiveness of this process.

More specifically, we conduct a preprocessing step to filter out less valuable data by means of the reward model. For each prompt, the model generates $K$ responses. If the variance in rewards for these responses is smaller than a margin $\epsilon$, it indicates that the prompt offers limited opportunity for exploration and improvement against the current model, and is therefore removed from the training dataset. Consequently, the final training dataset includes only those instances where significant variation and diversity are observed in the responses generated by the model.

\vpara{Proximal Policy Optimization (PPO)} is an online reinforcement learning framework for policy improvement. 
During PPO, we seek to optimize the policy model by maximizing the cumulative reward. 
\begin{equation}
    \mathrm{argmax}_{\pi_{\theta}} \mathbb{E}_{x\sim \mathcal{D},y_g\sim \pi_{\theta}}[r_\phi(x,y_g)]
\end{equation}
where $\pi_{\theta}$ is the policy model and $\theta$ is its learnable weights.  
In each training iteration, the policy model first generates responses $\{y_{i}\}$ for a set of prompt $\{x_i\}$. Then the reward model evaluates and assigns scores to each response, leading to $\{r(x_i,y_i)\}$, which is used to direct the gradient descent update of the policy model. 

Since the reward reflects approximate human preference and could be inaccurate, to avoid the policy diverging too far from the initial SFT model $\pi_0$, a penalty term based on KL divergence is usually added as part of the reward as regularization and also helps maintain training stability. 
\begin{equation}
    \begin{split}
    r(x, y_g) &= r_\phi(x,y_g) - \beta D_{KL}(\pi_{\theta}(y_g|x) || \pi_{0}(y_g|x)) \\
    &= r_\phi(x, y_g) - \beta \log \frac{\pi_{\theta}(y_g|x)}{\pi_{0}(y_g|x)}
    \end{split}
\end{equation}

\begin{wrapfigure}{r}{0pt}
\centering
\begin{minipage}[t]{0.5\columnwidth}
\vspace{-3mm}
\includegraphics[width=\linewidth]{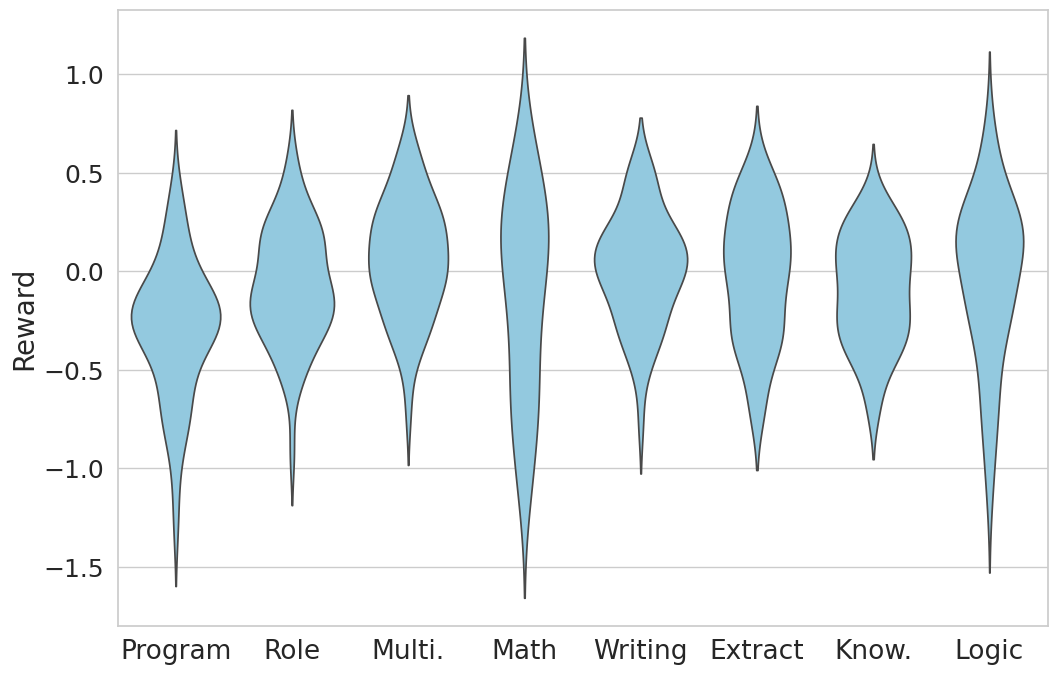}
\vspace{-6mm}
\caption{Reward distributions on different tasks.}
\label{fig:reward_distribution}
\vspace{-3mm}
\end{minipage}
\end{wrapfigure}

\vpara{Reward Bias Reduction} In the previous section, we make efforts to mitigate potential biases in reward modeling, such as length bias. Despite these efforts, it is observed that biases could not be entirely eliminated. These biases predominantly stem from two factors:
1) Value instability. The reward model utilizes a pairwise training loss; however, it does not impose explicit constraints regarding the numerical values assigned to responses in relation to their quality. Consequently, this lack of definitive guidance necessitates the implementation of reward or advantage normalization in the preliminary stages of PPO training to maintain stability.
2) Task/sample bias. The objective to enhance the general capabilities of the policy model necessitates training data that encompasses a wide range of tasks and prompts, including creative writing, mathematical reasoning, and role-playing. This diversity leads to the reward model assigning highly variable rewards across different prompts and tasks. This variability is attributed to differences in response style, length, and the Discriminativeness of the reward model, among other factors.
Figure ~\ref{fig:reward_distribution} illustrates the significant disparities in reward scores both within a single task and across different tasks. Consequently, a response that receives a high reward cannot be reliably deemed as high-quality without considering the influence of its associated prompt and task.

To overcome the above drawbacks, we introduce a reference baseline to avoid the variability from absolute value. The intuition is that the target is to enable the policy model to generate a better response than the reference one as much as possible, no matter its absolute reward. This corresponds to the initial objective of human preference and reward model training which excels at pairwise comparison rather than pointwise scoring.
To be specific, given a prompt $x$, we generate a response $y_{ref}$ during pre-processing as the reference response. During policy training, we regard the reward difference between the generated response $y$ and the reference response $y_{ref}$ as the optimization target,
\begin{equation}
    r(x, y) = \left(r_\phi(x, y_g) - r_\phi(x, y_{ref})\right) - \beta D_{KL}\left(\pi_{\theta}(y_g|x) || \pi_{0}(y|x)\right)
\end{equation}
At the warm-up training stage, $y$ and $y_{ref}$ are drawn from almost the same distribution, and thus $r(x,y)$ starts from near zero, which also helps stabilize the training even without other normalization tricks. ~\cite{wu2023pairwise} adopts a similar idea to replace absolute feedback with relative feedback

\vpara{Capability Forgetting}
The RLHF is expected to increase the response quality and better align them with human preferences. However, as a post-training step subsequent to SFT, we also observed unexpected behaviors in the policy after the RLHF stage. The model shows a reduced capability in handling specific scenarios, including wrong self-identification and invalid JSON format, and shifts a little far from the original model unexpectedly.
For example, the model is trained to claim itself as "ChatGLM" after SFT yet it would describe itself as "an AI assistant" after RLHF if asked \textit{"who are you?"}.
This behavior could be 
attributed to the problem of difference among data distributions or the inability of the reward model in such nuanced details. 
The problem of capability forgetting (or shifting) gets little attention in previous studies. 
But it is critical for online deployment,  especially for addressing diverse edge cases in real-world scenarios. 
For example, output in JSON format is crucial for tasks involving semantic analysis or answer extraction in practical applications. The emergence of capability forgetting can be attributed to the variance in data distribution between SFT and RLHF training, which encapsulates different facets of a model's capabilities.

To overcome the issue of capability forgetting, we propose to incorporate an extra supervised next-token-prediction loss as anadditionaln regularization besides the KL divergence, when performing reward maximization. This is intended to preserve the pre-existing abilities of the SFT model. The high-level idea is to encourage the model's outputs to align with human preferences through RLHF and leveraging next-token prediction to capture more granular signals.
To be more specific, in each training step, the training involves two aspects of data and objective: 1) RLHF training with prompts and model-generated responses for preference optimization,
$\mathcal{L}_r=-\sum_{x \sim \mathcal{D}} r(x,y_g)$; 
2) next-token-prediction with a small amount of human-annotated (prompt, response) pairs $\mathcal{D}_{S}$, which are specific to particular tasks and serve as supervised regularizations,
\begin{equation}
\mathcal{L} = \mathcal{L}_{r} - \beta \cdot \sum_{(x,y)\sim \mathcal{D}_{S}} \log \pi_{\theta}(y|x)
\end{equation}

\vpara{A simplified alternative: Direct Preference Optimization (DPO)} PPO has shown special advantages as an online algorithm, but it also brings additional technical difficulties in efficient parallel training in large language models as it involves both gradient decent update and response sampling during training.
Direct Preference Optimization (DPO)~\cite{rafailov2023direct} has been proven to be a simple yet effective alternative to PPO. DPO proposes to directly learn from annotated preference data without training a reward model. In our practice, we also implement DPO for comparison and find it is more flexible yet very effective. Different from the implementation reported in the original paper, we still train a reward model and regard it as a discriminator to help construct training data pairs with varied reward scores$(x, y_w, y_l)$. This way separates training data construction and data annotation and is thus more scalable. In DPO, the policy is trained to optimize the following objective:
\begin{equation}
    \mathcal{L}_{r} = -\sum_{(x, y_w, y_l)\sim \mathcal{D}}\log \sigma(\beta \log \frac{\pi_{\theta}(y_w|x)}{\pi_0(y_w|x)} - \beta \log \frac{\pi_{\theta}(y_l|x)}{\pi_0(y_l|x)} )
\end{equation}

\para{Parallel Training}
To efficiently train large models, the policy is trained with a combination of data parallelism and model parallelism~\cite{rasley2020deepspeed,narayanan2021efficient}. Model parallelism is further divided into tensor parallelism, which involves the division of tensors, and pipeline parallelism, which distributes different layers of a network across multiple GPUs.
Pipeline parallel can effectively reduce the communication overhead between GPUs during training with large batch size.
Typically, both the pretraining phase and supervised finetuning leverage a three-dimensional parallelism strategy—combining data, tensor, and pipeline parallelism—to optimize efficiency.
In the context of DPO, which relies solely on pre-generated responses, its training aligns with that of SFT, allowing for the application of three-dimensional parallelism.
But things are different for PPO, which involves the fusion of both gradient descent and inference within its training process.
Each generation comprises hundreds or thousands of forward operations and would incur a number of unexpected bubbles in the pipeline parallel and significantly slow down the generation. 
The generation generally accounts for more than 80\% training time, and it is crucial to avoid these bubbles for efficiency.
Consequently, in PPO training, we adopt a two-dimensional parallelism approach, combining data and tensor parallelism, to circumvent these bottlenecks.
There are public implementations based on fully-sharded data parallel~\cite{yao2023deepspeed, havrilla-etal-2023-trlx} but their efficiency is not satisfactory when finetuning full parameters of large-scale models due to the heavy communication cost. 

\para{RLHF v.s. Rejection Sampling}
Previous studies~\cite{touvron2023llama} have shown the effectiveness of Rejection-Sampling Fine-Tuning (RFT) as a post-SFT step in enhancing the performance of model alignment. RFT involves generating multiple outputs from the policy model and selecting the best candidates with the reward as the gold standard for further finetuning. The main difference between RFT and DPO is that DPO utilizes both positive and negative pairs for optimization, whereas RFT focuses on the positive instances. Both methods are constrained to pre-generated responses for training. PPO, an online algorithm, dynamically updates the model weights based on responses generated from the latest policy during training, which would generally produce better results.

\begin{table*}[ht]
    \centering
    \small
    \caption{Results of automatic evaluation on AlignBench (Chinese alignment) rated by gpt-4-0613. \textit{Prof. Know.}: professional knowledge. \textit{Fund. Lang.}: fundamental language. \textit{Logic.}: logical reasoning.}
    \renewcommand\tabcolsep{4pt}
    \begin{tabular}{l|cccccccc|c}
    \toprule[1.2pt]
         & \begin{tabular}{@{}c@{}}Prof. \\ Know.\end{tabular} & \begin{tabular}{@{}c@{}}Advanced \\ Chinese\end{tabular} & \begin{tabular}{@{}c@{}}Fund. \\ Lang.\end{tabular} & Math & Writing & OpenQA & \begin{tabular}{@{}c@{}}Role \\ Play\end{tabular} & \begin{tabular}{@{}c@{}}Logic.\end{tabular} & Overall \\
    \midrule
    GPT-3.5-turbo-0613$^1$ & 6.77 & 5.81 & 6.71 & 5.68 & 7.03 & 7.29 & 7.28 & 5.02 & 6.08 \\
    GPT-4-0613$^1$ & 7.94 & 6.93 & 7.81 & 7.56 & 7.93 & 7.42 & 7.51 & 7.37 & 7.53 \\
    GPT-4-1106-preview$^1$ & 8.65 & 7.33 & 7.99 & 7.80 & 8.67 & 8.61 & 8.47 & 7.66 & 8.01 \\
    GLM-4 (0116) & 8.52 & 8.19 & 7.29 & 7.26 & 8.27 & 8.50 & 8.53 & 7.30 & 7.75 \\
    \midrule
     Chinese-LLaMa-2-7B-Chat$^1$ & 4.13 & 4.26 & 4.31 & 2.29 & 4.63 & 4.50 & 4.91 & 3.07 & 3.57 \\ 
     Qwen-7B-Chat$^1$ & 5.66 & 5.74 & 6.40  & 3.62 & 6.31 & 6.26 & 6.19 & 3.83 & 4.91 \\
     ChatGLM3-6B-SFT & 4.80 & 5.17 & 5.21 & 2.71 & 6.72 & 6.55 & 6.34 & 4.02 & 4.58 \\
     ChatGLM3-6B-DPO & 5.64 & 4.60 & 4.94 & 3.41 & 6.85 & 7.29 & 6.66 & 3.81 & 4.80 \\
     ChatGLM3-6B-PPO & 5.93 & 5.08 & 5.28 & 3.57 & 7.13 & 6.84 &  6.16 & 3.90 & 4.90 \\
    \midrule
     DeepSeek-67B-Chat & 7.37 & 6.52 & 7.12 & 5.71 &	7.20 & 7.58 & 6.91 & 5.79 & 6.43 \\
     DeepSeek-67B-Chat-DPO	 & 7.71 & 7.47 & 7.29 & 6.13 & 7.51 & 7.82 & 7.83 & 5.41 & 6.69	\\
     ChatGLM-32B-SFT$^2$     & 7.47 & 6.62 & 6.53 & 5.87 & 7.45 & 7.13 & 7.38 & 5.42  & 6.37 \\
     ChatGLM-32B-DPO & 7.27 & 6.41  & 6.87 & 6.12 & 7.97 & 8.34 & 8.17 & 5.60  & 6.68 \\
     ChatGLM-32B-PPO & 7.27 &  6.97 & 7.06 & 5.85 & 7.92 & 8.18 & 7.96 &  6.02 & 6.75 \\
    \bottomrule[1.2pt]
    \end{tabular}
     \begin{tablenotes}
        \footnotesize
        \item $1$ The results are from the AlignBench paper, which employs varying generation temperatures for different tasks. Notably, all ChatGLM models produce responses in a greedy setting. In our experiments, the performance in the greedy setting is typically slightly lower than in the original setting but is stable.
        \item $2$ The version of ChatGLM is ChatGLM-32B-2310.
    \end{tablenotes}
    \label{tab:alignbench}
\end{table*}

\section{Experiments}


\subsection{Experimental Setup}
We conduct experiments based on ChatGLM~\cite{zeng2023glm,Du2021GLMGL} and test the effectiveness of 
DPO and PPO as the post-training step after SFT, as RLHF is always implemented based on an SFT model. 
We mainly focus on the improvement of the RLHF-aligned model over SFT model via both automatic evaluation and human evaluation.  

\vpara{Training Details} For PPO training, the policy and critics models share the same learning rate 1e-6 and are trained for around 600 to 800 iterations. Neither reward nor advantage normalization is adopted as reference rewards already lead to a stable training process in our experiments. At inference time, the policy top-p is set to 0.9 to generate diverse responses. 
In this section, we report the results on ChatGLM-6B and ChatGLM-32B models.


\subsection{Automatic Evaluation}
\vpara{Setup} To evaluate the alignment performance on tasks in Chinese, we assessed the ability of our trained language models across various fields using the AlignBench test set~\cite{liu2023alignbench}, a benchmark for open-ended questions. AlignBench comprises 8 main categories and 36 subcategories of tasks, covering a total of 683 questions. 
Alongside each question, AlignBench provides a reference answer and evaluation criteria, facilitating the automatic assessment of response quality by LLM-as-judge. We follow the original setting and use GPT-4 as the judge. The evaluation framework assembles the evaluation principles, the question, the model-generated response, and a reference answer as the input prompt to the LLM judge. The LLM-judge then performs the intended analytical evaluation and assigns a score ranging from 1 to 10 to the response.

We use the official code repository of AlignBench, available on GitHub\footnote{\url{https://github.com/THUDM/AlignBench}}. We generated responses to all questions with a greedy strategy to ensure the stability of the results and reduce variance, which is a little different from the original setting that sets different temperatures for different tasks.

\vpara{Results} The overall results are illustrated in Table~\ref{tab:alignbench}. We compare with ChatGPT~\cite{achiam2023gpt} and various open-source models~\cite{bi2024deepseek,bai2023qwen,touvron2023llama}. The results of the ChatGLM-32B show that RLHF methods, including both DPO and PPO, substantially improve the SFT model across a wide range of tasks in Alignbench. Notably, tasks such as Writing, OpenQA, and Role-Play exhibit more significant improvements. This suggests that the reward model is particularly suited to tasks related to creative writing, whereas it demonstrates limitations in tasks requiring advanced reasoning abilities, such as math and logic. This observation aligns with the intuitive understanding that human preferences tend to lean more toward stylistic and formatting elements rather than deep critical thinking skills.
Furthermore, PPO demonstrates a slight edge over DPO, with an average improvement margin of approximately 0.07. This outcome aligns with expectations, considering that PPO's more intricate design demands significantly greater resources during training compared to DPO.


\begin{table}[]
    \centering
    \caption{Results of human evaluation of ChatGLM-32B on internal test set. \textmd{We report the proportion of wins and ties of the PPO and SFT model on different tasks.}}
    \begin{tabular}{c|ccc|c}
    \toprule[1.2pt]
         &  PPO win & Tie & SFT win & $\Delta$(PPO-SFT) \\
    \midrule
        Multilingual & 0.32 & 0.44 & 0.24 & 0.08\\
        Creative Writing & 0.40 & 0.52 & 0.08 & 0.32 \\
        Role-play & 0.42 & 0.42 & 0.16 &  0.26 \\
        Knowledge & 0.36 & 0.50 & 0.14 &  0.22 \\
        Semantic Extraction & 0.30 & 0.50 & 0.20 & 0.10  \\
        Math. & 0.16 & 0.70 & 0.14 & 0.02 \\
        Programming & 0.22 & 0.72 & 0.06 & 0.16  \\
        Logic & 0.20 & 0.64 & 0.16 & 0.04 \\
    \midrule
        Overall & 0.30 & 0.55 & 0.15 & 0.15 \\
    \bottomrule[1.2pt]
    \end{tabular}
    \vspace{-2mm}
    \label{tab:human_eval}
\end{table}

\subsection{Human Evaluation}
\vpara{Setup.} 
In addition to automatic evaluations, we also incorporate human evaluations to assess the effectiveness of RLHF. Human evaluation is important because they're less likely to be biased as long as the rules are clearly defined. We collect a diverse human evaluation dataset, consisting of 400 instructions in Chinese, spanning an extensive array of tasks such as creative writing, logical reasoning, semantic analysis, language comprehension, and mathematics. Each subject area was represented by 50 distinct samples. 
The evaluation process employed a pairwise comparison method, where human annotators were tasked with selecting the more suitable response from a pair, with the option to declare a tie. And we have set very lenient requirements for determining whether two responses can be labeled as a tie, so that the \textit{win} could denote a clear advantage of one response over the other.
To minimize variability, each sample is annotated by two evaluators. In this part, we mainly compare ChatGLM-32b-SFT and ChatGLM-32b-PPO, as PPO has shown an edge over DPO in automatic evaluation.

\vpara{Results} Table~\ref{tab:human_eval} shows the results of human evaluation and we report the win-rate of two models. The results reveal that the PPO model maintains a distinct advantage over the SFT model, with a win-rate of 30\% compared to 15\%. Regarding task-specific performance, the findings align with the observations from the automatic evaluations. Notably, the improvement in mathematical reasoning tasks are minimal and barely significant, whereas tasks related to writing exhibit considerable improvements. Moreover, an unexpected yet significant advancement was observed in programming tasks. However, based on our experience, it remains challenging for the reward model to accurately identify errors within code snippets. The programming instructions primarily focus on practical guidance, such as \textit{how to build an Anaconda in linux}. This diverges from conventional code-evaluation benchmarks like HumanEval~\cite{chen2021evaluating}, requiring the model to solve practical programming problems. Additionally, a high tie rate of 55\% was recorded, aligning with our anticipations.

\hide{
\vpara{Case Study} Through human evaluation, we also found that RLHF can improve ChatGLM to better understand and follow human instructions.
which is not reflected in Table~\ref{tab:alignbench} and ~\ref{tab:human_eval}. 
Figure~\ref{fig:case_study} shows a representative case, in which the PPO model can better capture the indication of "romantic" and the conversation between two people.
This aligns with the notion that the reward model can provide both positive and negative signals, discouraging undesired behaviors while promoting desired ones.
}

\subsection{Evaluation on the Reward Model}

We conduct experiments to evaluate the accuracy of the reward model in predicting human preferences. Specifically, we create a test set comprising pairwise comparisons of tasks from human evaluation, each accompanied by carefully-examined human annotations. 
At inference time, the reward model is capable of producing a scalar value for an individual response without the need for a paired sample. 
We utilize the reward model to predict the reward for each response and assess whether it assigns higher reward scores to the responses preferred by humans.

The results are presented in Table~\ref{tab:reward_model}. We also include the accuracy of Llama2 reward model as a baseline for comparison. Despite ChatGLM and Llama2 being evaluated using distinct internal datasets, their results are still comparable as both models can reflect the alignment between human preferences and the predictions of the reward model. The data indicates that ChatGLM-32B surpasses ChatGLM-6B in terms of accuracy, as expected. However, the highest accuracy achieved by ChatGLM-32B is 65\%, which is on par with that of Llama2-70B. The marginal accuracy gain with increased model scale suggests a limit to the benefits of scaling. Nonetheless, the fact that the reward model can guide the training of RLHF algorithms with approximately 65\% accuracy in mirroring human judgment is noteworthy and deserves further exploration.

\begin{table}[]
    \centering
    \caption{Consistency between reward model and human preference on the internal test set. \textmd{The results of Llama2 are from~\cite{touvron2023llama}. ChatGLM and Llama2 are not evaluated on the same test set. But the results could still be comparable in how accurately the reward model predicts human preferences. We report Accuracy. }}
    \begin{tabular}{l|cc}
    \toprule[1.2pt]
         &  Training Acc & Test Acc \\
    \midrule
        Llama2-7B &  - & $\sim$0.61\\
        Llama2-13B & - & $\sim$0.61 \\
        Llama2-70B & - & $\sim$0.64 \\
        ChatGLM-6B & 0.64 & 0.59 \\
        ChatGLM-32B & 0.68 & 0.65 \\
    \bottomrule[1.2pt]
    \end{tabular}
    \label{tab:reward_model}
\end{table}

\begin{table}[]
    \centering
    \caption{Response length of different models (Tokens) . \textmd{Both DPO and PPO significantly increase the response length.}}
    \begin{tabular}{lll}
    \toprule[1.2pt]
         & ChatGLM-6B & ChatGLM-32B \\
    \midrule
        SFT model & 255.3 & 305.6 \\
       \qquad  + DPO & 427.5 & 410.9 (+32.6\%) \\
       \qquad  + PPO & 237.4 & 379.6 (+22.9\%) \\
    \bottomrule[1.2pt]
    \end{tabular}
    \vspace{-2mm}
    \label{tab:length}
\end{table}

\begin{figure}[ht]
    \centering
    \begin{minipage}[t]{0.35\textwidth}
        \includegraphics[width=\textwidth]{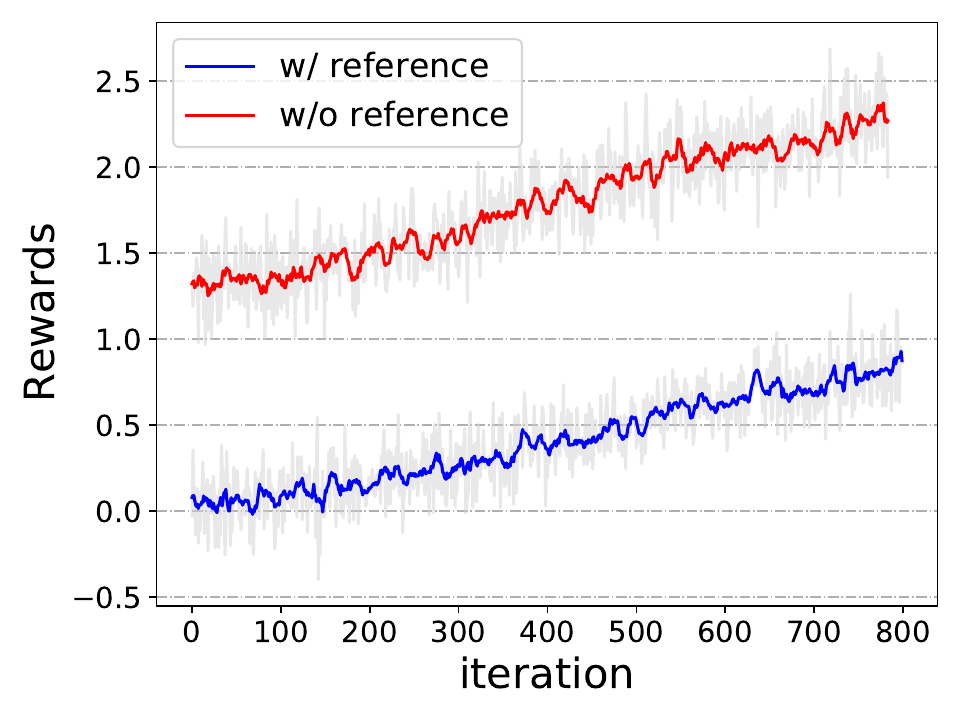}
    \end{minipage}
    \begin{minipage}[t]{0.35\textwidth}
        \includegraphics[width=\textwidth]{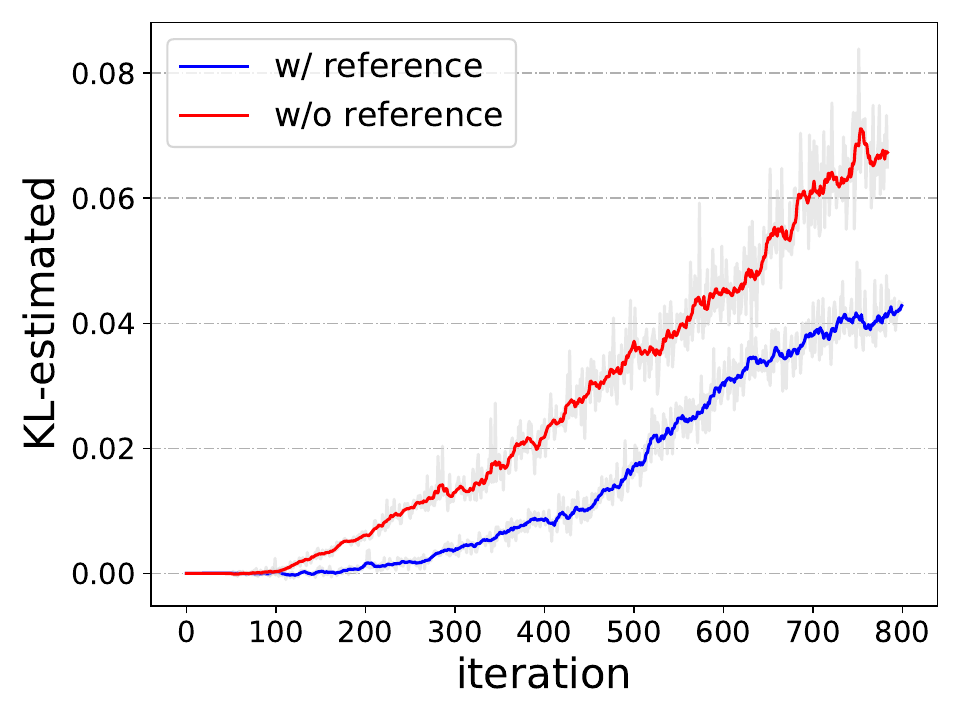}
    \end{minipage}
    \caption{Ablation on the effects of reference reward. \textmd{The two figures show the reward and estimated-KL during the training process of ChatGLM-32B-PPO.}}
    \vspace{-2mm}
    \label{fig:reference_reward}
\end{figure}

\subsection{Ablation Study}

\vpara{Response Length} In automatic evaluation, GPT-4 serves as the automated evaluator for assessing response quality. Previous study~\cite{zheng2023judging} suggests that GPT-4 is inclined to assign higher scores to lengthier responses, although the response may not imply better quality and instead denotes redundancy. To explore this aspect, we examine the lengths of responses generated by our models and report the results in Table~\ref{tab:length}. It is observed that all models produced responses that surpassed the SFT model in length. Notably, RLHF methods exhibit a remarkable increase in response length, i.e., 110 tokens in DPO and 75 tokens in PPO. Therefore, the improvement in automatic evaluation scores for creative writing tasks might be partly attributed to the increased response lengths. Moreover, the comparison between PPO and DPO reveals that despite the fact that PPO can generate shorter responses on average, it outperforms DPO in terms of performance. This indicates that PPO is capable of generating higher-quality responses with less redundancy.

\vpara{Effects of Reference Rewards} In our implementation of PPO, we employ a reference reward by deducting the reward of a baseline response. To evaluate the effectiveness of this approach, we compare PPO trained with (\textit{w/}) and without (\textit{w/o}) the reference reward, under the same hyper-parameter setting. Figure \ref{fig:reference_reward} illustrates the progression of reward and estimated KL-divergence throughout the training process. Remarkably, both experiments show similar patterns in reward improvement. However, the starting point of the reward value for the \textit{w/ ref} scenario is near zero and this contributes to the stabilization of the training process, which is also observed in the training loss curve. Moreover, the increase in estimated KL-divergence for the \textit{w/o ref} scenario is much lower, thereby drifting less far from the original SFT model, as expected.

\hide{
\begin{figure}
    \centering
    \includegraphics[width=0.45\textwidth]{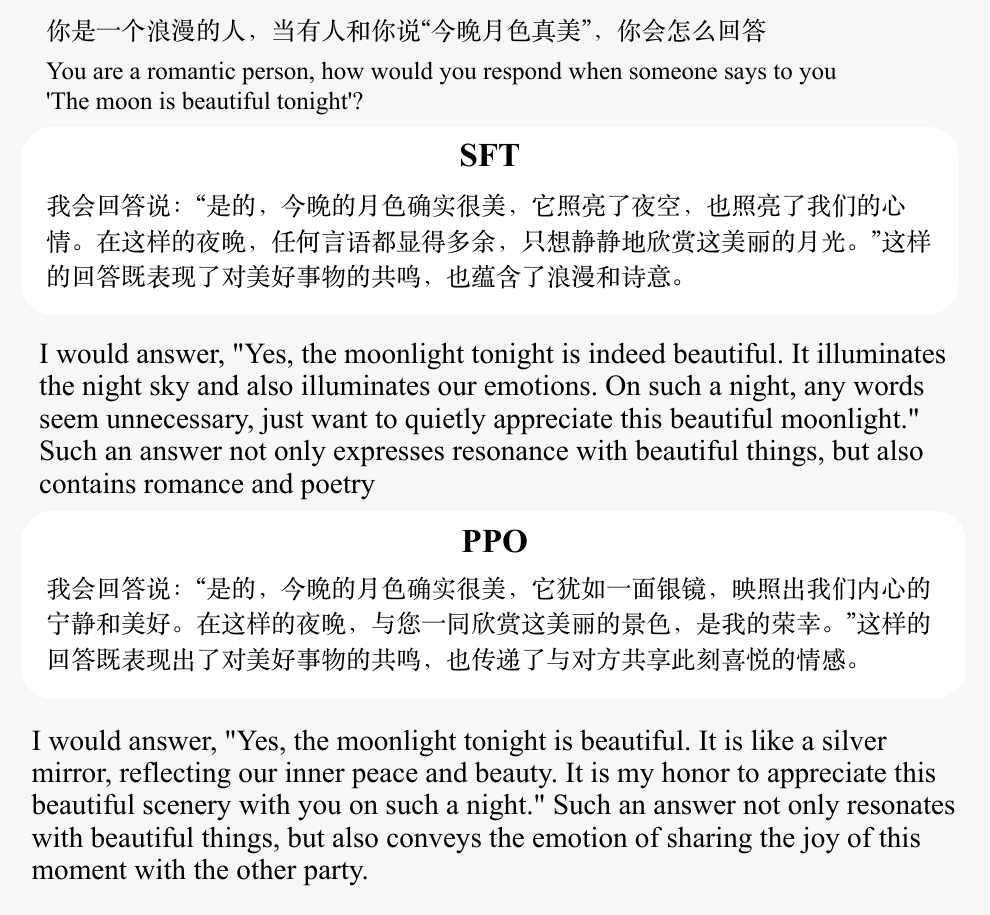}
    \vspace{-1mm}
    \caption{Case study of human evaluation.}
    \label{fig:case_study}
\end{figure}
}


\section{Conclusion}
This paper introduces ChatGLM-RLHF, a reinforcement learning pipeline devised to enhance the alignment of ChatGLM with human preference.
We provide an in-depth explanation of the three main components in ChatGLM-RLHF, data collection, reward model training, and policy model training.
Additionally, we discuss the challenges encountered during the practices and the corresponding solutions, such as how to mitigate reward bias and variance, implement model parallelism, and avoid catastrophic forgetting. Experimental results indicate a significant improvement of ChatGLM-RLHF over ChatGLM-SFT, marking an average of 15\% more wins in alignment with human preference, which underscores the efficacy of the proposed pipeline.

\bibliographystyle{ACM-Reference-Format}
\bibliography{reference}

\end{document}